# A Parsimonious Setup for Streamflow Forecasting using CNN-LSTM

Sudan Pokharel[1] and Tirthankar Roy [1]*

[1]Civil and Environmental Engineering, University of Nebraska-Lincoln, Lincoln, USA

*Corresponding Author: Tirthankar Roy (roy@unl.edu)

**Highlights**

- A CNN-LSTM model was developed for time series forecasting of streamflow in Nebraska by combining CNN for spatial data and LSTM for sequence data.
- A substantial improvement was observed for 66% of the basins for this model compared to the standalone LSTM.
- This superior performance was achieved just by using gridded precipitation and 2-m temperature as exogenous inputs.

**Abstract**

Significant strides have been made in advancing streamflow predictions, notably with the introduction of cutting-edge machine-learning models. Predominantly, Long Short-Term Memories (LSTMs) and Convolution Neural Networks (CNNs) have been widely employed in this domain. While LSTMs are applicable in both rainfall-runoff and time series settings, CNN-LSTMs have primarily been utilized in rainfall-runoff scenarios. In this study, we extend the application of CNN-LSTMs to time series settings, leveraging lagged streamflow data in conjunction with precipitation and temperature data to predict streamflow. Our results show a substantial improvement in predictive performance in 21 out of 32 HUC8 basins in Nebraska, showcasing noteworthy increases in the Kling-Gupta Efficiency (KGE) values. These results highlight the effectiveness of CNN-LSTMs in time series settings, particularly for spatiotemporal hydrological modeling, for more accurate and robust streamflow predictions.

## 1. Introduction

Streamflow prediction is one of the most challenging tasks in water resources management and is vital for various sectors, e.g. agriculture, disaster mitigation, planning and management of water resources. (Le et al., 2021; Pokharel et al., 2023). It is an integrated practice involving complex nonlinear physical processes that operate at multiple temporal and spatial scales within a watershed (Dehghani et al., 2023; Girihagama et al., 2022). Accurate streamflow prediction allows for advanced warning of potential floods, giving people in flood-prone areas time to prepare and implement flood control measures. Due to the proliferation of high-resolution and remote sensing datasets at high spatial and temporal resolutions, the use of deep learning (DL) algorithms has gained traction in the field of hydrology, especially in streamflow prediction (Dehghani et al., 2023; Ding et al., 2020; Hunt et al., 2022; Kratzert et al., 2018a, 2019; Pokharel et al., 2023; Vatanchi et al., 2023; Xiang et al., 2020; Yan et al., 2019).

Streamflow is typically modeled in two different ways. The first method involves time series modeling, where we incorporate both lagged predictors and lagged target values over a specific time period (look-back window) to make predictions. In contrast, the second method, i.e., rainfall-runoff modeling, focuses solely on the predictors up to a given time (t) without considering the lagged target values.

The Long Short-Term Memory (LSTM) (Hochreiter & Schmidhuber, 1997) model has gained traction and is widely used in streamflow prediction due to its ability to store and regulate information over time. It is well-suited for time series modeling because it can effectively remember long-term dependencies and handle sequential data (Bengio et al., 1994). In streamflow prediction, where the data often exhibits temporal patterns, LSTM can learn from historical information, including lagged predictors and target values, making them suitable for streamflow prediction (Khandelwal et al., 2020; Majeske et al., 2022).

Convolution Neural Networks (CNNs) are DL networks widely used for image and video recognition tasks. They are well known for their ability to learn spatial patterns within gridded data, such as images/videos, which is why this DL network is suitable for processing gridded spatial data in water resources. CNNs are a feedforward network like ANN. However, CNNs are more advanced than ANN with convolution and pooling layers, which help automatically extract the essential features from the input layer (Shu et al., 2021). CNNs can be 1-D, 2-D, and 3-D, depending on the shape of the input layer. In 1-D CNN, the inputs are vectors, whereas in 2-D CNN, the inputs are in gridded/matrix form. The majority of the studies in hydrology involving

CNN utilized the 1-D version of it (Chong et al., 2020; Haidar & Verma, 2018; Hussain et al., 2020), although a few studies applied 2-D CNN (Chen et al., 2021; Huang et al., 2020).

CNN-LSTM is a hybrid neural network that combines the strengths of both CNN and LSTM. It is a deep learning architecture designed for sequence prediction problems with spatial inputs, like images or videos (Donahue et al., 2017). The CNN model is used for feature extraction, while the LSTM model is used for interpreting the features across timesteps. Only a few studies in hydrology have applied CNN-2-D-LSTM in comparison to the widely-used LSTM model (Anderson & Radić, 2022; Li et al., 2023; Ng et al., 2023). Anderson & Radić (2022) used CNN-LSTM in rainfall-runoff settings using spatial data to predict streamflow at 226 stream gauges across southwestern Canada. They used gridded climate reanalysis data to train and predict streamflow and demonstrated the efficiency and suitability of the CNN-LSTM model for spatiotemporal hydrological modeling. Li et al. (2023) proposed a CNN-LSTM in rainfall-runoff settings that introduces actual evaporation as an additional training target and demonstrates the value of spatial information in hydrological modeling. Mohammed & Corzo (2024) applied CNN-LSTM in regional rainfall runoff settings to simultaneously predict daily streamflow in 86 catchments across the US, adding value by using spatial signals in the model. However, to the author's knowledge, no studies have investigated using CNN-LSTM in time-series settings. The contribution of this study is as follows.

1. Applying and testing LSTM and CNN-LSTM models for 32 Nebraska HUC8 basins in time-series settings.

2. Comparing the efficiency of those models in predicting the streamflow for Nebraska HUC8 basins.

3. Investigating the added value of using CNN-LSTM in time series settings for streamflow prediction.

## 2. Study Area

In this study, we used 32 HUC8 basins in Nebraska, which have at least 20 years of streamflow data available (Figure 1). Diverse physical and geographical features characterize these basins, including a range of landscapes such as the Great Plains, the Sand Hills, and the Ogallala

Aquifer. The area of the basins ranged from 1,366 to 13,352 km$^2$, with each exhibiting unique hydrological characteristics, such as different soil types, land-use patterns, and water availability.

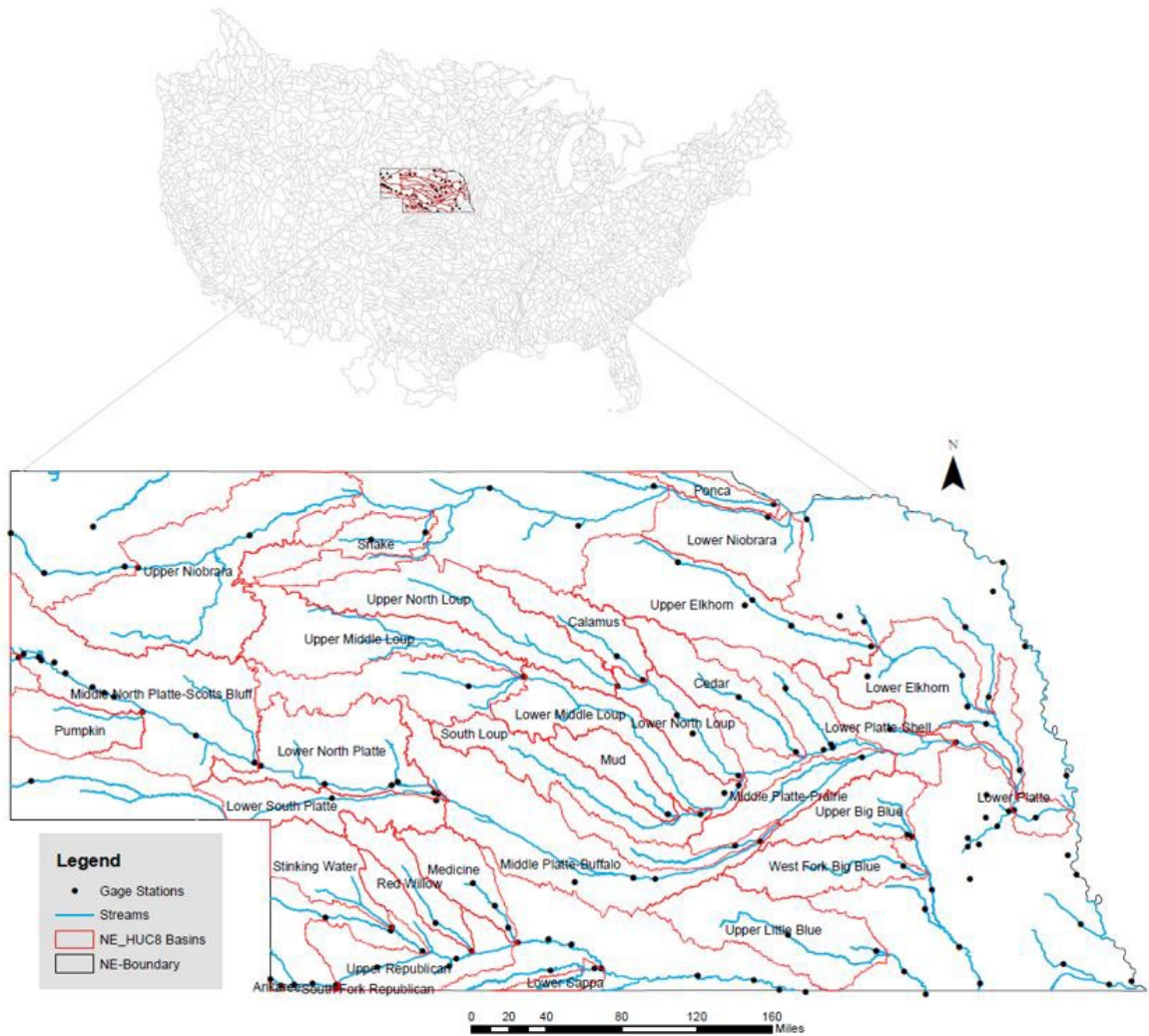

**Figure 1:** 32 HUC8 basins in Nebraska, as considered in this study.

3. **Data**

In this study, we used data from three sources. Precipitation and 2-meter temperature gridded data were obtained from ERA5-Land (9 km x 9 km) (Muñoz-Sabater et al., 2021), which is the fifth-generation atmospheric reanalysis from the European Centre for Medium-Range Weather Forecasts (ECMWF). Streamflow data were collected from the United States Geological Survey (USGS) and the Nebraska Department of Natural Resources (NeDNR). The datasets from the ERA5-Land were aggregated on daily timescales by taking an average over the basin for LSTM since LSTM models cannot process gridded data. For the CNN-LSTM model, gridded data was used.

We know that hydrology is complex and has many underlying physics that govern the streamflow. However, we are only taking into consideration precipitation and temperature because these are the two principal meteorological inputs for hydrological modeling (Anderson & Radić, 2022; Essou et al., 2016) and to showcase the enhanced improvement in streamflow prediction by using CNN-LSTM in time-series settings. Thus, taking a parsimonious approach, we tried to build a simpler model with few predictor variables to compare the efficacy of models.

## 4. Methods

### 4.1. Convolution Neural Networks (CNN)

Convolution Neural Networks (CNNs) (LeCun et al., 1998) is a deep learning model designed to process spatial data with a grid pattern like images and videos. CNN consists of three building-block layers: convolution, pooling, and fully connected (Figure 2) (Khosravi et al., 2020; Yamashita et al., 2018). The convolution layer applies a filter to the input tensor to extract features by detecting the spatial pattern. The convolution layer slides filters, also known as kernels, over the input tensor to compute dot products between the filter and input at each location, resulting in feature maps. The pooling layer reduces the spatial size of the feature map by down-sampling it, reducing the number of parameters and computations in the network. Max pooling is commonly used as it takes the maximum activation value in each filter region. The fully-connected layer connects every neuron in one layer to every neuron in the next layer, takes the previous layer's output, and produces the network's final output.

In CNN, strides are the number of pixels the filter moves each time during the convolution operation. Padding refers to adding pixels of zero around the input image borders. Padding is done so that when the filters slide over the edge pixels, there is meaningful information for it to process rather than an empty space. The output size depends on the input dimensions, strides, padding, and filter size.

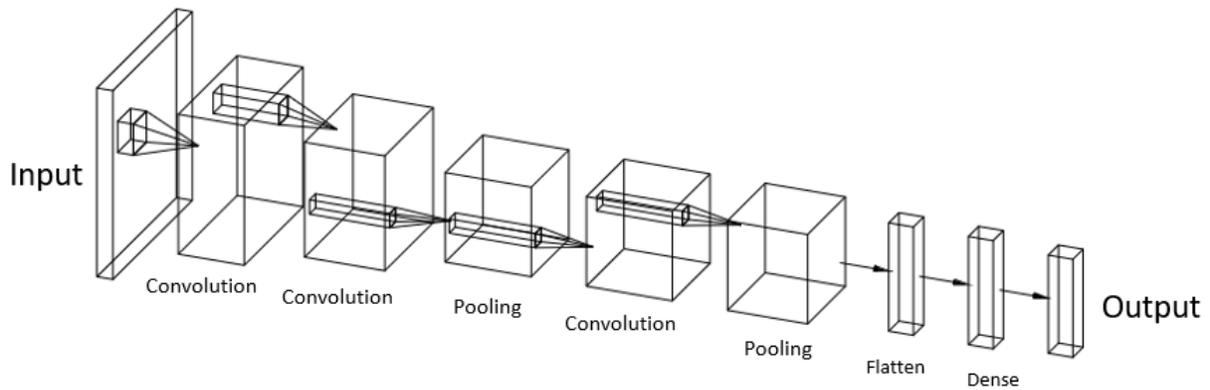

**Figure 2:** Architecture of Convolution Neural Network (Drew with the help of an online tool created by LeNail (2019).

### 4.2. Long Short-Term Memory (LSTM)

LSTM (Hochreiter & Schmidhuber, 1997) is an evolution form of the RNN, which was developed to address the problem of vanishing and exploding gradients in RNN. LSTM consists of three main gates: the forget, the input, and the output (Figure 3). As the name suggests, the forget gate determines which information should be discarded based on the previous hidden state ($h_{t-1}$) and current input data ($X_t$). The input gate adds the new information to the network's long-term memory given the previous hidden state ($h_{t-1}$) and current input data ($X_t$). The output gate determines the next hidden state ($h_t$) and is used to make the final predictions.

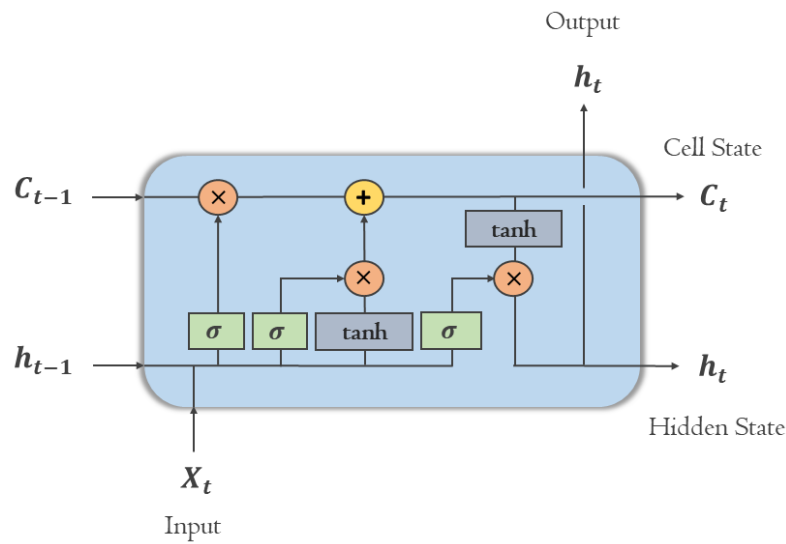

**Figure 3:** An Architecture of the LSTM model

## 5. Methodology

### 5.1. Time-Series LSTM

In this experiment, we utilized the LSTM neural network for time series modeling. The lagged values of predictors (Precipitation, 2-meter temperature) and the lagged streamflow values were input to calculate streamflow at time t. The gridded datasets underwent basin averaging as LSTM models cannot directly handle spatial data in its raw form. They were aggregated to form coherent input sequences for the LSTM model. The objective was to forecast streamflow at time t by examining t-n lagged predictors along with the target value.

### 5.2. Time-Series CNN-LSTM

In this experiment, we developed an individual sequential CNN-LSTM model for streamflow forecasting. The goal was to leverage the complementary strengths of CNN for extracting spatial features and LSTM for modeling temporal dynamics (Figure 4). Analogous to video frames, we treated the daily gridded meteorological input data (precipitation and 2-meter temperature) as image-like frames. Each frame had a defined spatial structure spanning geographic coordinates (latitude and longitude), similar to image rows and columns. The frames also comprised precipitation and temperature channels, like RGB color channels. Stacked over the model's 6-month or 182-day input sequence, these daily weather "images" formed an input "video" for precipitation and temperature over the catchment area.

The strategy for applying a CNN-LSTM model for time series analysis involved representing lagged streamflow values as channels. The original 1-D discharge time series underwent a transformation into a 2-D tensor by duplicating values across the spatial grid. This integration allowed the streamflow data to be seamlessly concatenated with meteorology grids along the channel dimension. The model's input was comprised of a six-month-long video encompassing 182 frames. Each frame was equipped with three channels representing precipitation, 2-meter temperature, and streamflow. These frames underwent a series of time-distributed convolution and pooling layers, as depicted in Figure 2. The CNN component efficiently extracted crucial features from each frame, generating 182 vectors that encapsulate spatially learned features for a specific day. Subsequently, these feature vectors traversed through the LSTM layer to capture temporal dependencies. The dense layer further mapped the discharge values derived from the LSTM output.

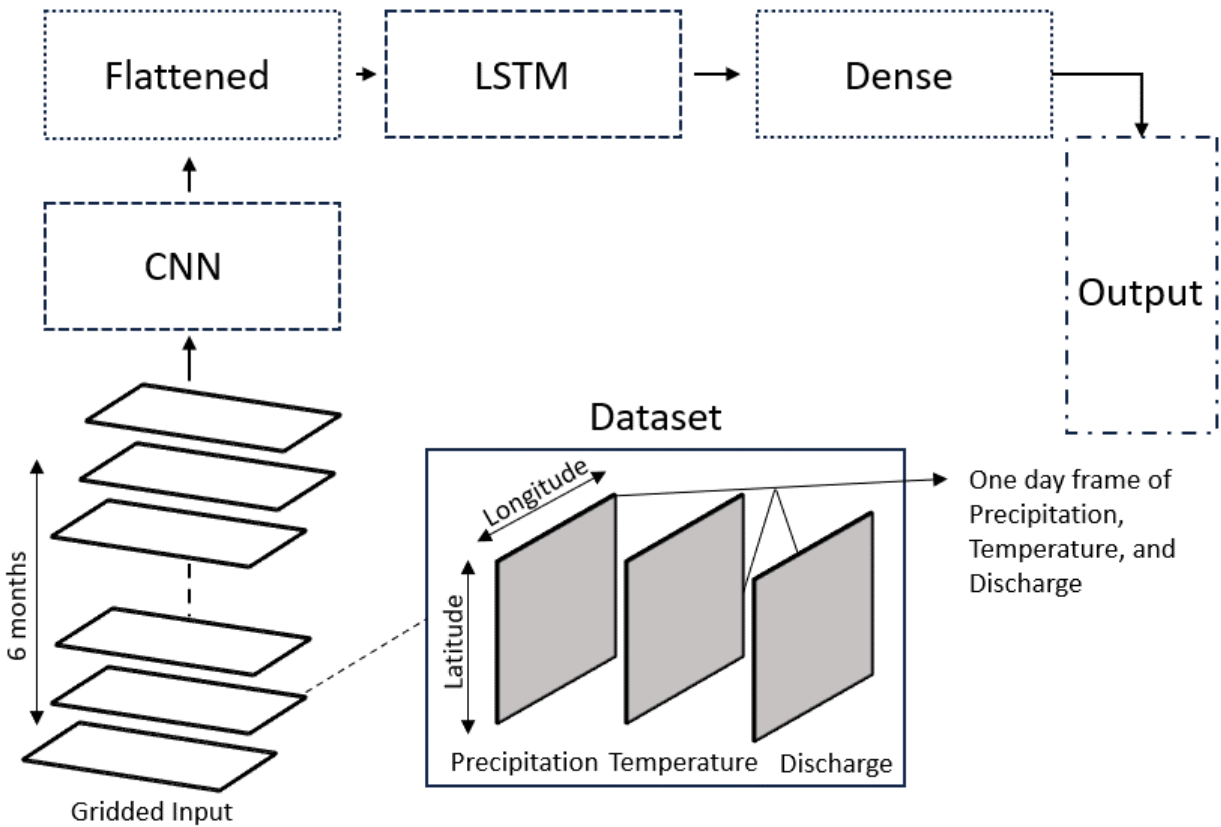

**Figure 4:** Overview of CNN-LSTM model architecture in a time series setting.

## 6. Training of models

The initial step in our experiment involved transforming the datasets into the requisite format for model training. Then, it was split into training and testing sets in a 70:30 ratio. The validation split was 0.3, meaning 30% of the training data was used for validation. The datasets were normalized using the standard scaler method, subtracting the mean and dividing by the standard deviation. Only the training period datasets were used to calculate the mean and standard deviation. For consistency and to see the effectiveness of models, we used the same parameter across all the experiments carried out. The activation function, batch size, optimizer, and dropout were kept the same for all the experiments. We used *Relu* as the activation function, *Adam* as the optimizer, and the epoch and batch size were kept 100 and 50 respectively. Dropout, kernel initializer (he_uniform), and gradient norm were used to avoid overfitting, vanishing, and exploding gradient problems. The number of LSTM layers and neurons was kept the same for the LSTM and CNN-LSTM models. Table 1 shows the CNN-LSTM internal architecture. The use of 1x1 convolutional filters followed by an increase to 3x3 filters provided a way to manage model complexity, control the number of feature maps, and reduce computational requirements in CNNs, leading to more efficient and effective network architectures (Anderson & Radić, 2022; Lin et al., 2013; Szegedy et al., 2014, 2015). We also used a single LSTM layer with 80 neurons, as it was found in the previous studies that using a single LSTM layer with more neurons performed better than multiple layers with fewer neurons (Anderson & Radić, 2022; Kratzert et al., 2018b).

**Table 1:** Description of CNN-LSTM model layers.

| Layers | Type of Layers | Description |
|---|---|---|
| 0 | Convolution | 32 filters, 1x1 |
| 1 | Convolution | 16 filters, 3x3 |
| 2 | Max Pooling2D | 1x1 |
| 3 | Convolution | 32 filters, 1x1 |
| 4 | Max Pooling2D | 1x1 |
| 5 | Dropout | 0.3 |
| 6 | LSTM | 80 |
| 7 | Dense | 1 |

All the basins were trained separately using the same configuration as discussed above. Finally, all the model performances were evaluated using Kling-Gupta Efficiency (Gupta et al., 2009).

## 7. Evaluation metrics

This study used Kling-Gupta efficiency (KGE) (Gupta et al., 2009) to assess the models' predictive performance. KGE is a widely used metric in hydrology that integrates three common types of errors in a model, i.e., bias, variability, and timing (correlation), and is written as:

$$KGE = 1 - \sqrt{(r-1)^2 + (\beta-1)^2 + (\gamma-1)^2}$$

where $r$ is the linear correlation coefficient between the observations and simulations. $\gamma$ is the ratio of the coefficients of variation, and $\beta$ is the ratio of the means. KGE can range from -∞ to 1, with a value of 1 indicating perfect correspondence between simulations and observations.

## 8. Results and discussion

### 8.1. Comparison of individual CNN-LSTM and LSTM performance

In this section, we compare the performance of LSTM and CNN-LSTM in time-series settings across 32 Nebraska HUC8 basins. The model's predictions were used to evaluate the model's ability to simulate streamflow during the testing phase. Out of 32 river basins analyzed, the CNN-LSTM model performed better in 62% (21 basins) compared to a standard LSTM model for streamflow prediction. The maximum increment in model performance (KGE metric) was from 0.39 to 0.91 as we moved from LSTM to CNN-LSTM, with an overall median KGE improvement from 0.76 to 0.78 (Figure 6). The CNN-LSTM model demonstrated unique prediction capabilities, possibly attributed to its utilization of the CNN component. This contribution is likely associated with CNN's ability to learn spatial features that enhance prediction accuracy.

In 11 basins; however, the CNN-LSTM model underperformed compared to the LSTM model. These basins were characterized by a high density of flood control structures, potentially affecting the CNN-LSTM model's ability to match the spatial signal with streamflow, especially in regions with extensive cropland and irrigation. Flood control structures like dams and reservoirs disrupt natural precipitation-discharge relationships in several ways. They regulate flow rates, redirect water, retain and slowly release floods, and dampen extreme signals (Gunnell et al., 2019; Hovis et al., 2021). Additionally, their operating policies change over time, reducing the relevance of older training data (Sung et al., 2018). While structures help manage regional water resources, they alter historical signals that data-driven models rely on.

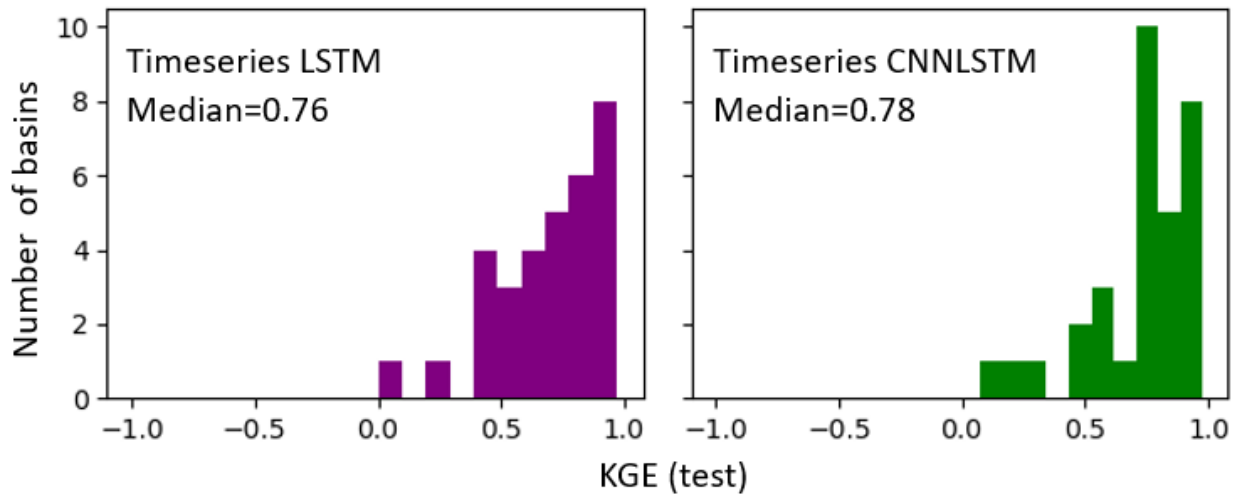

**Figure 6:** Distribution of KGE values in LSTM and CNN-LSTM in time-series settings.

In considering the applicability of the CNN-LSTM model in regions with established streamflow gauges, its usefulness remains evident. This model proves advantageous due to the persistent challenges associated with achieving accurate streamflow measurements. These include the influence of backwater from ice or debris, such as log jams, and algae and aquatic growth in the stream (Kiang et al., 2018). Additionally, sediment movement and the potential malfunction of measurement equipment further increase the chance of inaccurate streamflow measurement. The precision and accuracy of these measurements are of high importance, as they directly influence the reliability of hydrologic models, and any inaccuracies directly translate into uncertainties within hydrologic models (Dos Santos et al., 2021). Furthermore, accurate streamflow is critical for various sectors, such as water resources planning and management, agriculture water management, disaster mitigation, and infrastructure safety.

## 9. Conclusion

This study presented a new approach for implementing a CNN-LSTM model using lagged streamflow data in a time-series setting. The performance of the CNN-LSTM and LSTM models was compared across 32 basins in Nebraska. The key finding is that the CNN-LSTM model, trained on 1-D streamflow data and gridded precipitation and temperature inputs, demonstrates improved performance over the LSTM model in 21 of the 32 test basins. These results suggest the feasibility and potential benefits of applying CNN-LSTM (time-series setting) models for streamflow prediction tasks. The model's ability to capture intricate spatial patterns and signals between predictors and targets surpasses the capabilities of standard LSTM recurrent networks. However, the study also uncovered degradation in CNN-LSTM performance as compared to the standard LSTM in some basins that are characterized by high streamflow magnitudes and extensive irrigated croplands.

The value of this study lies in demonstrating a new approach of using CNN-LSTM for streamflow prediction and its potential benefits. While this study used the same set of hyperparameters to showcase the efficacy of CNN-LSTM in time-series settings throughout all basins, it should be noted that other variables such as solar radiation, wind, vapor pressure, and some basin static attributes such as slope, soil, and land characteristics could result in even better model performances.

**Conflict of Interest**

The authors declare no conflict of interest.

**Funding:**

The authors acknowledge funding from the University of Nebraska Collaboration Initiative Grant (Award Number 32370).